\documentclass[runningheads]{llncs}
\usepackage{graphicx}

\usepackage{tikz}
\usepackage{comment}
\usepackage{amsmath,amssymb} 
\usepackage{color}
\usepackage{booktabs}
\usepackage{multirow}
\usepackage{orcidlink}
\usepackage[accsupp]{axessibility}  

\begin{document}
\pagestyle{headings}
\mainmatter
\def\ECCVSubNumber{1185}  

\title{Tracking Objects as Pixel-wise Distributions}


\titlerunning{P3AFormer}
%
\author{Zelin Zhao\inst{1}\orcidlink{0000-0002-2638-0414}\thanks{The work was done when Zelin Zhao took internship at SmartMore.} \and
Ze Wu\inst{2}\orcidlink{0000-0002-6084-0939} \and
Yueqing Zhuang\inst{2}\orcidlink{0000-0001-8614-3536} \and Boxun Li\inst{2}\orcidlink{0000-0002-6132-7547} \and Jiaya Jia\inst{1,3}\orcidlink{0000-0002-1246-553X}}
\authorrunning{Z. Zhao et al.}
%
\institute{The Chinese University of Hong Kong \and
MEGVII Technology \and SmartMore}
\maketitle
\newcommand{\leo}[1]{{\color{blue}{#1}}}
\definecolor{airforceblue}{rgb}{0.36, 0.54, 0.66}
\newcommand{\TODO}[1]{{\color{airforceblue}{TODO: #1}}}
\begin{abstract}
Multi-object tracking (MOT) requires detecting and associating objects through frames. Unlike tracking via detected bounding boxes or center points, we propose tracking objects as pixel-wise distributions. We instantiate this idea on a transformer-based architecture named P3AFormer, with pixel-wise propagation, prediction, and association. P3AFormer propagates pixel-wise features guided by flow information to pass messages between frames. Further, P3AFormer adopts a meta-architecture to produce multi-scale object feature maps. During inference, a pixel-wise association procedure is proposed to recover object connections through frames based on the pixel-wise prediction. P3AFormer yields 81.2\% in terms of MOTA on the MOT17 benchmark -- highest among all transformer networks to reach 80\% MOTA in literature. P3AFormer also outperforms state-of-the-arts on the MOT20 and KITTI benchmarks. The code is at https://github.com/dvlab-research/\\ECCV22-P3AFormer-Tracking-Objects-as-Pixel-wise-Distributions.

\end{abstract}
\section{Introduction}
Multi-Object Tracking (MOT) is a long-standing challenging problem in computer vision, which aims to predict the trajectories of objects in a video. Prior work investigates the tracking paradigms \cite{fairmot,2016SimpleOnline,2017deepsort,centerTrack}, optimizes the association procedures \cite{LearnableGraphMatching,bytetrack} and learns the motion models \cite{ObjectPerformance,DetectionmotionModeling}. Recently, with the powerful transformers deployed in image classification~\cite{vit,SwinTransformer} and object detection~\cite{DETR,deformabledetr,SwinTransformer}, concurrent work applies transformers to multi-object tracking~\cite{Transcenter,TransMOT,MOTR,Transtrack}. Albeit the promising results, we note that the power of the transformer still has much room to explore.

\begin{figure}[t]
\centering
\includegraphics[width=\textwidth]{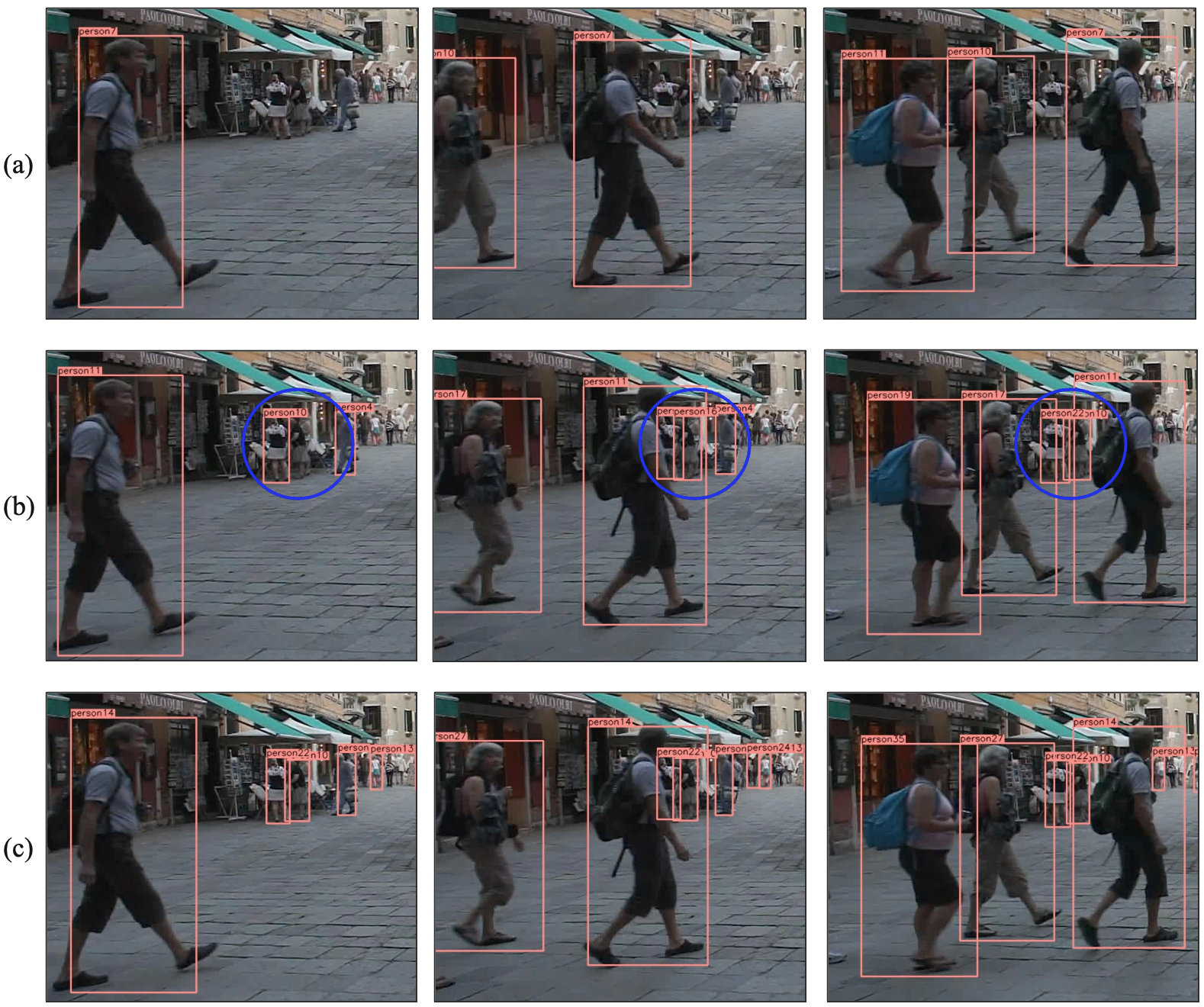}
\caption{Comparison of concurrent transformer-based MOT approaches on the \texttt{MOT17-val} dataset. Each row corresponds to one method. (a) MOTR~\cite{MOTR} occasionally fails to detect the small objects. (b) TransCenter~\cite{Transcenter} has a lot of ID-switches (indicated by blue circles) when the small objects are occluded. (c) Our proposed P3AFormer can robustly track the small objects under occlusion.}
\label{fig:teaser}
\end{figure}

As shown in Figure~\ref{fig:teaser}, current transformer-based MOT architectures MOTR~\cite{MOTR} and TransCenter~\cite{Transcenter} still face challenges in detecting small objects and handling occlusions. Representing objects via pixel-wise distributions and conducting pixel-wise associations might alleviate these issues for the following reasons. First, pixel-wise information may help overcome occlusion based on low-level clues~\cite{densefusion,pvnet,DensePredictionVIT}. Moreover, recent transformer architecture demonstrates strong performance in pixel-wise prediction~\cite{SwinTransformer,maskformer,mask2former}. From another perspective, pixel-wise prediction preserves more low-confident details, which can improve tracking robustness \cite{bytetrack}.

We propose a transformer approach, called P3AFormer, to conduct pixel-wise propagation, prediction, and association. P3AFormer propagates information between frames via the dense feature propagation technique~\cite{flowGuidedFeatureAggregation}, exploiting context information to resist occlusion. To produce robust pixel-level distribution for each object, P3AFormer adopts the meta-architecture \cite{maskformer,mask2former}, which generates object proposals and object-centric heatmaps. Masked attention \cite{mask2former} is adopted to pursue localized tracking results and avoid background noise. Further, P3AFormer employs a multi-scale pixel-wise association procedure to recover object IDs robustly. Ablation studies demonstrate the effectiveness of these lines of improvement.

Besides these pixel-wise techniques, we consider a few whistles and bells during the training of P3AFormer. First, we use Hungarian matching~\cite{DETR} (different from direct mapping) to enhance detection accuracy. Second, inspired by the empirical findings of YOLOX~\cite{yolox}, we use strong data augmentation, namely mosaic augmentation, in the training procedure of P3AFormer. Further, P3AFormer preserves all low-confident predictions~\cite{bytetrack} to ensure strong association. 

We submit our results to the MOT17 test server and obtain 81.2\% MOTA and 78.1\% IDF1, outperforming all previous work. On the MOT20 benchmark, we report test accuracy of 78.1\% MOTA and 76.4\% IDF1, surpassing existing transformer-based work by a large margin. We further validate our approach on the KITTI dataset. It outperforms state-of-the-art methods. Besides, we validate the generalization of the pixel-wise techniques on other MOT frameworks and find that these pixel-wise techniques generalize well to other paradigms.
\section{Related Work}
\subsection{Transformer-based multiple-object tracking}
We first discuss concurrent transformer-based MOT approaches. TrackFormer~\cite{TrackFormer} applies the vanilla transformer to the MOT domain and progressively handles newly appeared tracks. TransTrack~\cite{Transtrack} and MOTR~\cite{MOTR} take similar approaches to update the track queries from frames. They explore a tracking-by-attention scheme, while we propose to track objects as pixel-wise distributions. TransCenter~\cite{Transcenter} conducts association via center offset prediction. It emphasizes a similar concept of dense query representation. However, the model design and tracking schemes are different from ours. TransMOT~\cite{TransMOT} is a recent architecture to augment the transformer with spatial-temporal graphs. It is noted that our P3AFormer does not use graph attention. We validate different paradigms and model components in experiments to support the design choices of our method.

\subsection{Conventional multi-object tracking}
The widely used MOT framework is tracking-by-detection~\cite{mot_review,2016SimpleOnline,2017deepsort,detect2track,MOTviaMultiCutandReid,fairmot,bytetrack,tractor}. DeepSORT~\cite{2017deepsort} leverages the bounding box overlap and appearance features from the neural network to associate bounding boxes predicted by an off-the-shelf detector~\cite{fasterrcnn}. Yang et al.~\cite{MOTviaMultiCutandReid} propose a graph-based formulation to link detected objects. Other work tries different formulations. For example, Yu et al.~\cite{rlMOT} formulate MOT into a decision-making problem in Markov Decision Processes (MDP). CenterTrack~\cite{centerTrack} tracks objects through predicted center offsets. Besides, it is also investigated to reduce post-processing overhead by joint detection and tracking~\cite{JDTByGNN,RetinaTrack,DEFT}. Another line of work~\cite{TrackMPNN,LearnableGraphMatching,GraphMOT} leverages graph neural networks to learn the temporal object relations.

\subsection{Transformer revolution}
Transformer architectures achieved great success in natural language processing (NLP)~\cite{AttentionIsAllYouNeed,bert}. Recently, transformer demonstrated strong performance in various vision tasks, such as image classification~\cite{vit,SwinTransformer,maxvit,v2x-vit}, object detection~\cite{DETR,deformabledetr}, segmentation~\cite{maskformer,mask2former,SegFormer}, 3d recognition~\cite{MVSTER} and pose estimation~\cite{MHFormer,Exploiting}. The seminal work~\cite{DETR} proposes a simple framework DETR for end-to-end object detection. MaskFormer~\cite{maskformer} utilizes a meta-architecture to generate pixel embeddings and object proposals via transformers jointly. Previous transformers use masks in attention to restrict attention region~\cite{AttentionIsAllYouNeed} or force the computation to be local~\cite{SwinTransformer,ProTo}.

\subsection{Video object detection}
Tracking by detection paradigm requires accurate object detection and robust feature learning from videos~\cite{wu2022end,huang2022rife}. Zhu et al.~\cite{flowGuidedFeatureAggregation} propose dense feature propagation to aggregate features from nearby frames. The follow-up work~\cite{HighPerformanceVOT} improves aggregation and keyframe scheduling. The MEGA model~\cite{MEGA} combines messages from different frames on local and global scales. These methods do not consider video object detection with transformers. TransVOD~\cite{TransVOD} proposes aggregating the transformer output queries from different frames via a temporal query encoder. TransVOD cannot be directly applied to our setting because it is not an online algorithm and does not make pixel-wise predictions.

\subsection{Pixel-wise techniques}
Pixel-wise techniques have been proven effective in various applications in computer vision. Dense fusion~\cite{densefusion} and pixel-wise voting network~\cite{pvnet,Obj6D} are proposed to overcome occlusions in the object pose estimation~\cite{fan2022object6D}. DPT~\cite{DensePredictionVIT} uses a dense prediction transformer for monocular depth estimation and semantic segmentation. Pyramid vision transformer~\cite{PyramidVisionTransformer} replaces convolutional neural networks by attention to dense prediction tasks. Yuan et al.~\cite{Hrformer} presents a high-resolution transformer for human pose estimation. Our P3AFormer, instead, explores the power of pixel-wise techniques in the MOT domain.
\begin{figure}[tp]
\centering
\includegraphics[width=\textwidth]{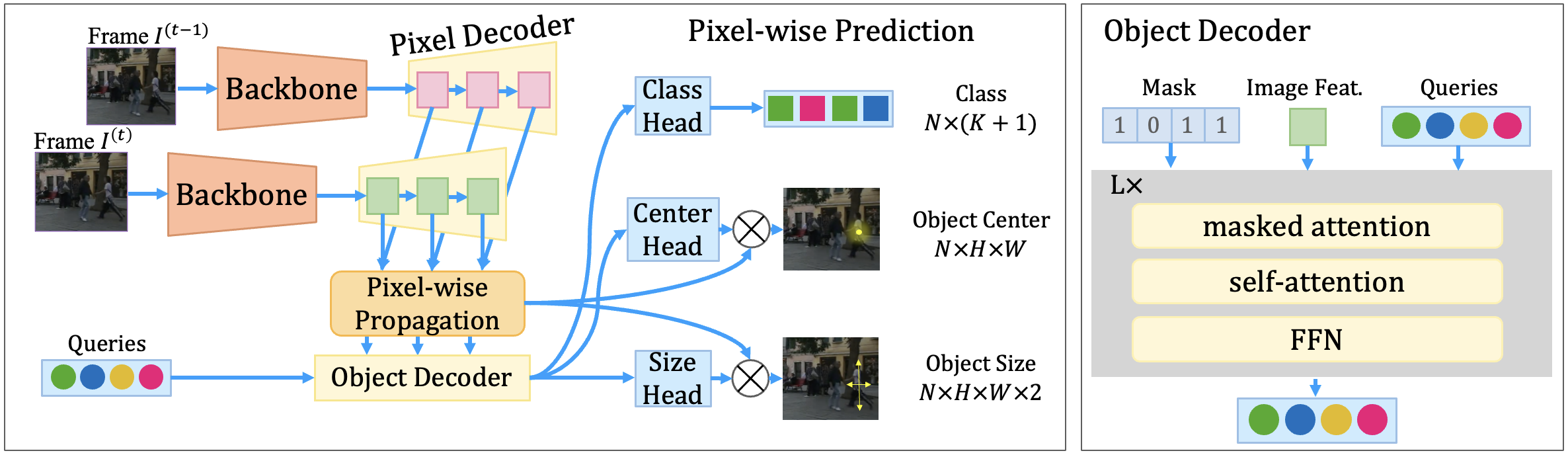}
\caption{Diagram of P3AFormer model. (\textbf{Left}) The backbone encodes the input images, and the pixel decoder produces pixel-level multi-frame feature embeddings. Then the object decoder predicts latent object features, which are passed through several MLP heads to produce class distribution and the pixel-wise representations for object center and size. (\textbf{Right}) The detailed structure of the object decoder. It uses masked attention, self-attention, and feed-forward networks (FFN) to update the query embedding. The add and normalization layers are omitted in this figure for simplicity.}
\label{fig:model_P3AFormer}
\end{figure}

\section{Pixel-wise Propagation, Prediction and Association}
\label{sec:p3aformer}
Different from tracking objects via bounding boxes~\cite{bytetrack,Transtrack,tractor} or as points~\cite{Transcenter,centerTrack}, we propose to track objects as pixel-wise distributions. Specifically, P3AFormer first extracts features from each single frame (Sec.~\ref{sec:feature_extraction_object_proposal}), summarizes features from different frames via pixel-wise feature propagation (Sec.~\ref{sec:pixel_level_feature_propagation}) and predicts pixel-wise object distributions via an object decoder (Sec.~\ref{sec:pixel_wise_predictions}). The training targets are listed in Sec.~\ref{sec:training_targets}. During inference, P3AFormer conducts pixel-wise association (Sec.~\ref{sec:dense_association}) to build tracks from object distributions.

\subsection{Single-frame feature extraction}
\label{sec:feature_extraction_object_proposal}
As shown on the top-left of Figure~\ref{fig:model_P3AFormer}, P3AFormer uses a backbone to generate latent features and a pixel decoder to produce pixel-wise heatmaps. The details are as follows.

\subsubsection{Backbone.} The input to the P3AFormer model is a set of continuous frames from a video. For simplicity and following previous work \cite{centerTrack,Transcenter}, we take two consecutive frames $\mathcal{I}^{(t-1)}$ and $\mathcal{I}^{(t)}$ as input. A backbone generates low-resolution features $\mathbf{F}^{(t)} \in \mathbb{R}^{d \times H_{\mathbf{F}}\times W_{\mathbf{F}}}$ from the input image $\mathcal{I}^{(t)}$, where $d$ is the feature dimension, $H_{\mathbf{F}}$ and $W_{\mathbf{F}}$ are the height and width of the extracted feature maps. Another backbone (sharing weight with the first backbone) extracts the previous-frame feature $\mathbf{F}^{(t-1)}$ from $\mathcal{I}^{(t-1)}$.

\subsubsection{Pixel decoder.} P3AFormer uses the pixel decoder~\cite{maskformer}, which is a transformer decoder, to up-sample the features $\mathbf{F}^{(t)}$ and generate per-pixel representation $\mathbf{P}^{(t)}_{l}$ where $l$ is the current decoding level. The pixel encoder is also applied to the previous-frame feature $\mathbf{F}^{(t-1)}$ to get the pixel-wise feature $\mathbf{P}^{(t-1)}_{l}$. In our work, we use a multi-scale deformable attention transformer \cite{deformabledetr} as the pixel decoder.

\subsection{Pixel-wise feature propagation}
\label{sec:pixel_level_feature_propagation}
Extracting temporal context from nearby frames is very important in MOT~\cite{2017deepsort,MOTR,centerTrack}. We use the pixel-wise flow-guided feature propagation~\cite{flowGuidedFeatureAggregation} to summarize features between frames (shown in the middle-left of Figure~\ref{fig:model_P3AFormer}). Formally, given a flow network $\Phi$~\cite{flownet}, the flow guidance can be represented as $\Phi(\mathcal{I}^{(t-1)}, \mathcal{I}^{(t)})$. After that, a bilinear warping function $\mathcal{W}$~\cite{flowGuidedFeatureAggregation} transforms the previous-frame feature to align with the current feature as
\begin{equation}
    \mathbf{P}_{l}^{(t-1)->(t)}=\mathcal{W}(\mathbf{P}_{l}^{(t-1)}, \Phi(\mathcal{I}^{(t-1)}, \mathcal{I}^{(t)})).
\end{equation}
We then compute the fused feature as
\begin{equation}
    \bar{\mathbf{P}_{l}}^{(t)}=\mathbf{P}_{l}^{(t)} + w^{(t-1)->(t)}\mathbf{P}_{l}^{(t-1)->(t)},
\end{equation}
where the weight $w^{(t-1)->(t)}$ is the pixel-wise cosine similarity~\cite{flowGuidedFeatureAggregation} between the warped feature $\mathbf{P}_{l}^{(t-1)->(t)}$ and the reference feature $\mathbf{P}_{l}^{(t)}$. The pixel-wise cosine similarity function is provided in the supplementary file for reference. The shape of $\mathbf{P}_l$ is denoted as $H_{P_l} \times W_{P_l} \times d$.

\subsection{Pixel-wise predictions} 
\label{sec:pixel_wise_predictions}
P3AFormer uses a transformer-based object decoder to generate object proposals. The object proposals are combined with the pixel-wise embeddings to get pixel-wise object distributions. As shown in the right part of Figure~\ref{fig:model_P3AFormer}, the object decoder follows the standard transformer~\cite{DETR,AttentionIsAllYouNeed}, which transforms $N$ learnable positional embeddings $\mathbf{Q}_l \in \mathbb{R}^{N \times d}$ using $L$ attention layers. 

Since an image from an MOT dataset~\cite{mot17,mot20} often involves a large number of small objects, local features around the objects are often more important than features at long range~\cite{fairmot}. Inspired by the recent discovery~\cite{mask2former} that masked attention~\cite{AttentionIsAllYouNeed,ProTo,mask2former} can promote localized feature learning, P3AFormer uses masked attention in each layer of the object decoder. The mask matrix $\mathbf{M}_{l}$ is initialized as an all-zero matrix, and it is determined by the center heatmaps at the previous level (presented in Eq.~\eqref{eq:mask_update}). The standard masked attention~\cite{AttentionIsAllYouNeed,ProTo,mask2former} can be denoted as
\begin{equation}
\label{eq:query_update}
    \mathbf{X}_{l}=\operatorname{softmax}\left(\mathbf{\mathbf{M}}_{l-1}+\mathbf{Q}_{l} \mathbf{K}_{l}^{\mathrm{T}}\right), \mathbf{V}_{l}+\mathbf{X}_{l-1},
\end{equation}
where $X_l$ is the hidden query feature at layer $l$ while the query feature is computed by a linear mapping from the hidden query feature of $\mathbf{Q}_{l} = f_Q(\mathbf{X}_{l-1})$. The key feature matrix $\mathbf{K}_l \in \mathbb{R}^{H_{P_l}W_{P_l} \times d}$ and the value feature matrix $\mathbf{V}_l \in \mathbb{R}^{H_{P_l}W_{P_l} \times d}$ are derived from the image feature as $\mathbf{K}_l = f_K(\bar{\mathbf{P}_{l}}), \mathbf{V}_l = f_V(\bar{\mathbf{P}_{l}})$. These functions $f_Q$, $f_K$ and $f_V$ are all linear transformations. As shown in the right part of  Figure~\ref{fig:model_P3AFormer}, after the masked attention, the hidden query feature passes through the standard self-attention and feed-forward networks. Please refer to \cite{AttentionIsAllYouNeed,maskformer} for these operator details.

At each level, the query embeddings are decoded via three MLP heads to get three embeddings corresponding to object class $\mathcal{E}^{cls}_l=\operatorname{MLP_{cls}}(\mathbf{Q}_l)\in \mathbb{R}^{N \times d}$, the object center $\mathcal{E}^{ctr}_l=\operatorname{MLP_{ctr}}(\mathbf{Q}_l)\in \mathbb{R}^{N \times d}$, and the size of the object $\mathcal{E}^{sz}_l=\operatorname{MLP_{sz}}(\mathbf{Q}_l)\in \mathbb{R}^{N \times d \times 2}$. The hidden dimension of the object size embedding is doubled because the bounding box size is represented in two dimensions (x- and y-axis). Given the full-image representation $\bar{\mathbf{P}_{l}}$ and the object-centric embeddings $\mathcal{E}^{ctr}_l$ and $\mathcal{E}^{sz}_l$, we compute the center heatmaps $c_l\in \mathbb{R}^{N \times H_{P_l} \times W_{P_l}}$ and size maps $s_l\in \mathbb{R}^{N \times H_{P_l} \times W_{P_l}\times 2}$ via dot products as
\begin{equation}
\begin{split}
c_l[i, h, w]=\operatorname{sigmoid}(\bar{\mathbf{P}_{l}}[h, w, :] \cdot \mathcal{E}^{ctr}_l[i, :]), \\
s_l[i, h, w, j]=\operatorname{sigmoid}(\bar{\mathbf{P}_{l}}[h, w, :] \cdot \mathcal{E}^{sz}_l[i, :, j]).
\end{split}
\end{equation}
After getting the center heatmaps, the attention mask corresponding to $i$-th object and position $(x, y)$ is updated as
\begin{equation}
\label{eq:mask_update}
    \mathbf{\mathbf{M}}_{l-1}(i, x, y) = \left\{\begin{array}{ll}0, & \text { if } c_{l-1}[i, x, y]>0.5, \\ -\infty, & \text { otherwise. }\end{array}\right.
\end{equation}
Such a mask restricts the attention to the local region around the object center, much benefiting the tracking performance (as shown in Sec.~\ref{sec:ablate_techniques}). 

\subsection{Training targets} 
\label{sec:training_targets}
P3AFormer leverages the bipartite Hungarian matching~\cite{DETR,deformabledetr} to match $N$ predicted objects to $K$ ground-true objects. During classification, the unmatched object classes are set to an additional category called ``no object" ($\varnothing$). Following MaskFormer~\cite{maskformer}, we adopt a pixel-wise metric (instead of bounding boxes) in the Hungarian matching. 

First, we construct the ground-true heatmap $h_l^i$ for an object via a Gaussian function where the Gaussian center is at the object center, and the Gaussian radius is proportional to the object size~\cite{centerTrack,Transcenter}. Given the predicted center heatmaps $\hat{h_l^i}$, class distribution $\hat{p_l^i}$ of 
the $i$th object, and the corresponding ground true center heatmaps $h_l^{i}$ and object class $c_l^{i}$, we compute the association cost between the prediction and the ground truth via the pixel-wise cost function of
\begin{equation}
\label{eq:dense_metric}
    \mathcal{L}_{pixel-wise} = \sum_l\sum_i\left(-\operatorname{log}\hat{p_l^i}(c_l^{i}) + \mathbf{1}_{c_{l}^{\mathrm{i}} \neq \varnothing} |\hat{h_l^i} - h_l^{i}|\right).
\end{equation}
P3AFormer further computes three losses given the matching between predictions and ground-true objects: (1) cross-entropy loss between the predicted and ground-true classes; (2) focal loss~\cite{centerTrack} between the predicted center heatmaps and ground-true center heatmaps; (3) size loss computed by the $L1$ loss between predicted and ground true size. The final loss is a weighted form of these three losses summarized for all levels.

\subsection{Pixel-wise association}
\label{sec:dense_association}

\begin{figure}[tp]
\centering
\includegraphics[width=\textwidth]{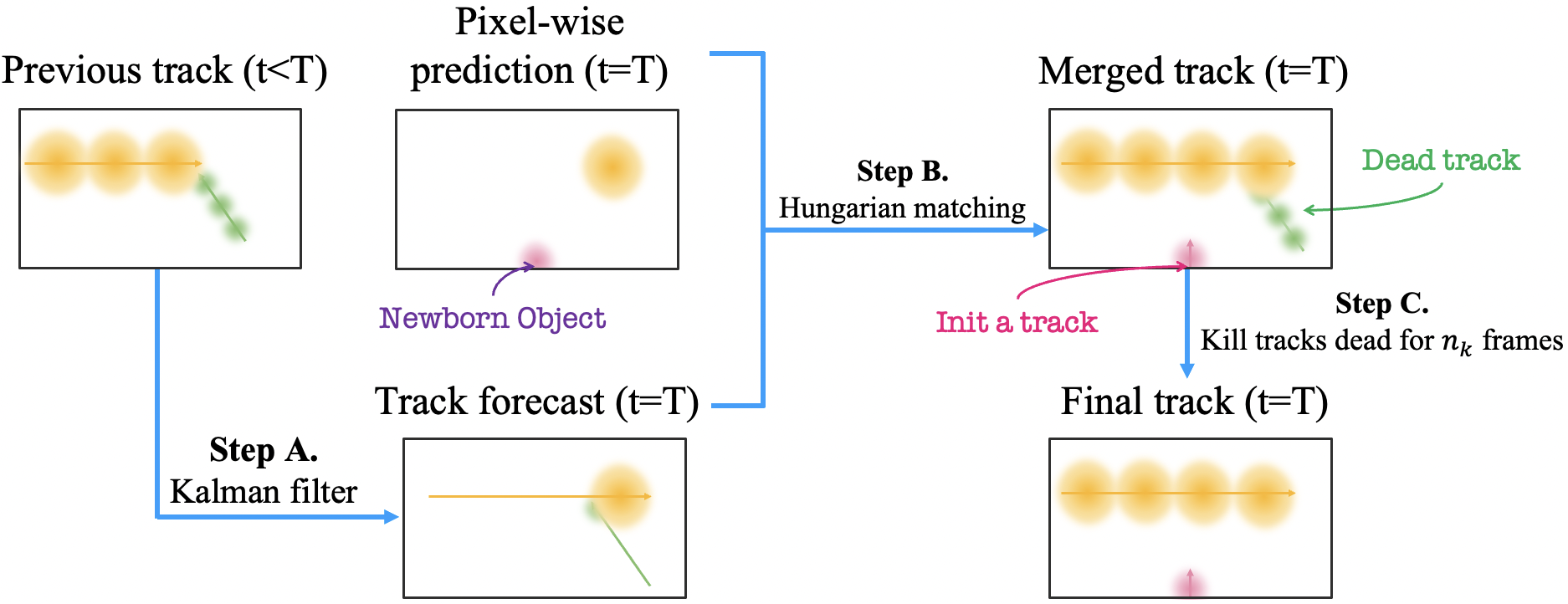}
\caption{Pixel-wise association scheme in P3AFormer. One object is represented as a pixel-wise distribution, denoted by spheres with the radial gradient change in this figure. We use one arrow and spheres on the arrow to denote a track. \textbf{Step A:} P3AFormer feeds the previous track into the Kalman filter~\cite{kalman_filter} to produce the track forecast. \textbf{Step B:} A Hungarian algorithm matches predictions and the track forecast based on a pixel-wise cost function (Eq.~\eqref{eq:dense_metric}). P3AFormer initializes a new track if a newborn object is detected. \textbf{Step C:} The dead tracks are removed when an object is occluded or moves out of the image, and the final track is obtained.}
\label{fig:pixel-wise-association}
\end{figure}

After representing objects as a pixel-wise distribution, P3AFormer adopts a pixel association procedure to recover object tracks across frames. This procedure is conducted from the first frame ($t=0$) to the current frame ($t=T$) in a frame-by-frame manner, which means P3AFormer is an online tracker. 

We sketch the association procedure at the timestep $t=T$ in Figure~\ref{fig:pixel-wise-association}. A track $\tau_k$ corresponds to an object with a unique ID $k$. We store into $\tau_k$ the bounding boxes \texttt{$\tau_k$.bbox} (recovered by the predicted center and size), the score \texttt{$\tau_k$.score} (the peak value in the center heatmap), predicted class \texttt{$\tau_k$.class} and the center heatmap \texttt{$\tau_k$.heatmap}. 

In \textbf{step A}, We use the Kalman Filter~\cite{kalman_filter,bytetrack} to predict the location of objects in the current frame ($t=T$) based on previous center locations ($t<T$). The heatmaps are translated along with the forecast movement of the object center via bilinear warping. \textbf{Step B} uses the Hungarian algorithm to match the pixel-wise prediction with the track forecast. P3AFormer only matches objects under the same category and omits the ``no-object" category. The association cost for Hungarian matching is the L1 distance between a track's forecast heatmap and an object's predicted heatmap. We accept a matching if the association cost between the matched track and prediction is smaller than a threshold $\eta_m$. 
A new track $\tau_{k'}$ would be initialized for an untracked object $k'$ if its confidence \texttt{$\tau_{k'}$.score} is larger than $\eta_s$. In \textbf{step C}, the dead tracks that are not matched with any prediction for $n_k$ frames are killed. All the above thresholds $\eta_m$, $\eta_s$, and $\eta_k$ are hyper-parameters detailed in Sec.~\ref{sec:implementation_details} and we provide more algorithm details in the supplementary file.

\section{Experiments}
P3AFormer accomplishes superior results to previous MOT approaches on three public benchmarks. We then ablate each component of the P3AFormer to demonstrate the effectiveness of the pixel-wise techniques. After that, we generalize the proposed pixel-wise techniques to other MOT frameworks.

\setlength{\tabcolsep}{4pt}
\begin{table}[t]
\begin{center}
\caption{Results on the MOT17 test set. We list transformer-based approaches on bottom of the table and others above. The numbers are from original papers released by authors. \textbf{Bold} numbers indicates the best model. We use $\dagger$ to denote unpublished work (prior to ECCV'22) and ``W\&B" represents whistles and bells.}
\label{tab:mot17TestResults}
\begin{tabular}{l|ccccccc}
\toprule
 Methods & \multicolumn{1}{c}{MOTA $\uparrow$} & \multicolumn{1}{c}{IDF1 $\uparrow$} & \multicolumn{1}{c}{MT $\uparrow$} & \multicolumn{1}{c}{ML $\downarrow$} & \multicolumn{1}{c}{FP $\downarrow$} & \multicolumn{1}{c}{FN $\downarrow$} & \multicolumn{1}{c}{IDSW $\downarrow$} \\ \midrule
FairMOT~\cite{fairmot} & $73.7$ & $72.3$ & $19.5$ & $36.6$ & 12201 & 248047 & 2072 \\
LSST17~\cite{LSST} & $54.7$ & 62.3 & $20.4$ & $40.1$ & 26091 & 228434 & \textbf{1243} \\
Tracktor v2~\cite{tractor} & $56.5$ & $55.1$ & $21.1$ & $35.3$ & \textbf{8866} & 235449 & 3763 \\
GMOT~\cite{GraphMOT} & $50.2$ & 47.0 & 19.3 & 32.7 & 29316 & 246200 & 5273 \\
CenterTrack~\cite{centerTrack} & $67.8$ & $64.7$ & $34.6$ & $24.6$ & 18498 & 160332 & 3039 \\
QuasiDense~\cite{quasidense} & 68.7 & $66.3$ & $40.6$ & $21.9$ & $26589$ & $146643$ & $3378$ \\
SiamMOT~\cite{siammot} & 65.9 & 63.3 & 34.6 & 23.9 & 14076 & 200672 & 2583 \\
PermaTrack~\cite{ObjectPerformance} & 73.8 & 68.9 & 43.8 & 17.2 & 28998 & 115104 & 3699 \\
CorrTracker~\cite{coortracker} & 76.5 & 73.6 & 47.6 & 12.7 & 29808 & 99510 & 3369 \\
ByteTrack$^{\dagger}$~\cite{bytetrack} & $80.3$ & $77.3$ & $53.2$ & $14.5$ & $25491$ & \textbf{83721} & $2196$ \\ \midrule
MOTR$^{\dagger}$~\cite{MOTR} & 73.4 & 68.6 & 42.9 & 19.1 & 27939 & 119589 & 2439 \\
TransTrack$^{\dagger}$~\cite{Transtrack} & 74.5 & 63.9 & 46.8 & \textbf{11.3} & 28323 & 112137 & 3663 \\
TransCenter$^{\dagger}$~\cite{Transcenter} & 73.2 & 62.2 & 40.8 & 18.5 & 23112 & 123738 & 4614 \\
TransMOT$^{\dagger}$~\cite{TransMOT} & 76.7 & 75.1 & 51.0 & 16.4 & 36231 & 93150 & 2346 \\
P3AFormer & 69.2 & 69.0 & 34.8 & 28.8 & 18621 & 152421 & 2769 \\
P3AFormer (+W\&B) & \textbf{81.2} & \textbf{78.1} & \textbf{54.5} & $13.2$ & $17281$ & 86861 & $1893$ \\
\bottomrule
\end{tabular}
\end{center}
\end{table}
\setlength{\tabcolsep}{1.4pt}

\setlength{\tabcolsep}{4pt}
\begin{table}[ht]
\begin{center}
\caption{Results on the MOT20 test set. The transformer-based approaches are the last three ones. We use \textbf{bold} numbers to indicate the best approach. We use $\dagger$ to denote unpublished work (prior to ECCV'22) and ``W\&B" represents whistles and bells.}
\label{tab:mot20TestResults}
\begin{tabular}{l|ccccccc}
\toprule
Methods & MOTA $\uparrow$ & IDF1 $\uparrow$ & MT $\uparrow$ & ML $\downarrow$ & FP $\downarrow$ & FN $\downarrow$ & IDs $\downarrow$ \\ \midrule
MLT~\cite{MLT} & 48.9 & 54.6 & 30.9 & 22.1 & 45660 & 216803 & 2187 \\
FairMOT~\cite{fairmot} & 61.8 & 67.3 & 68.8 & 7.6 & 103440 & 88901 & 5243 \\
CorrTracker~\cite{coortracker} & 65.2 & 69.1 & 66.4 & 8.9 & 79429 & 95855 & 5183 \\
Semi-TCL~\cite{semi-tcl} & 65.2 & 70.1 & 61.3 & 10.5 & 61209 & 114709 & 4139 \\
CSTrack~\cite{cstrack} & 66.6 & 68.6 & 50.4 & 15.5 & \textbf{25404} & 144358 & 3196 \\
GSDT~\cite{JDTByGNN} & 67.1 & 67.5 & 53.1 & 13.2 & 31913 & 135409 & 3131 \\
SiamMOT~\cite{siammot} & 67.1 & 69.1 & 49.0 & 16.3 & - & - & - \\
RelationTrack~\cite{relationtrack} & 67.2 & 70.5 & 62.2 & 8.9 & 61134 & 104597 & 4243 \\
SOTMOT~\cite{SOTMOT} & 68.6 & 71.4 & 64.9 & 9.7 & 57064 & 101154 & 4209 \\
ByteTrack~\cite{bytetrack} & 77.8 & 75.2 & 69.2 & 9.5 & 26249 & 87594 & \textbf{1223} \\ \midrule
TransTrack~\cite{Transtrack} & 65.0 & 59.4 & 50.1 & 13.4 & 27197 & 150197 & 3608 \\
TransCenter~\cite{Transcenter} & 61.9 & 50.4 & 49.4 & 15.5 & 45895 & 146347 & 4653 \\
P3AFormer & 60.3 & 56.2 & 50.4 & 13.5 & 43221 & 157564 & 4522 \\
P3AFormer (+W\&B) & \textbf{78.1} & \textbf{76.4} & \textbf{70.5} & \textbf{7.4} & 25413 & \textbf{86510} & 1332 \\ \bottomrule
\end{tabular}
\end{center}
\end{table}

\subsection{Datasets}
\subsubsection{MOT17~\cite{mot17}.} The MOT17 dataset is focused on multiple persons tracking in crowded scenes. It has 14 video sequences in total and seven sequences for testing. The MOT17 dataset is the most popular dataset to benchmark MOT approaches \cite{mot_review,bytetrack,2017deepsort,Transcenter,MOTR}. Following previous work \cite{bytetrack,centerTrack} during validation, we split the MOT17 datasets into two sets. We use the first split for training and the second for validation. We denote this validation dataset as \texttt{MOT17-val} for convenience. The best model selected during validation is trained on the full MOT17 dataset and is submitted to the test server under the ``private detection" setting. The main metrics are MOTA, IDF1, MT, ML, FP, FN, and IDSW, and we refer the readers to \cite{mot17} for details of these metrics. For MOT17, we add CrowdHuman~\cite{crowdhuman}, Cityperson~\cite{Citypersons}, and ETHZ~\cite{ethz} into the training sets following \cite{bytetrack,Transcenter}. When training on an image instead of a video with no neighboring frames, the P3AFormer model replicates it and takes two identical images as input.

\subsubsection{MOT20~\cite{mot20}.} The MOT20 dataset consists of eight new sequences in crowded scenes. We train on the MOT20 training split with the same hyper-parameters as the MOT17 dataset. We submit our inferred tracks to the public server of MOT20~\cite{mot20} under the ``private detection" protocol. The evaluation metrics are the same as MOT17.

\subsubsection{KITTI~\cite{kitti}.} The KITTI tracking benchmark contains annotations for eight different classes while only two classes ``car" and ``pedestrian" are evaluated~\cite{centerTrack,ObjectPerformance}. Twenty-one training sequences and 29 test sequences are presented in the KITTI benchmark. We split the training sequences into halves for training and validation following~\cite{centerTrack}. Besides the common metrics, KITTI uses an additional metric of HOTA~\cite{hota} to balance the effect of detection and association.

\subsection{Implementation details}
\label{sec:implementation_details}

\paragraph{Backbone.} We mainly use ResNet~\cite{resnet} and Swin-Transformer~\cite{SwinTransformer} as the backbone in P3AFormer. For ResNet, we use the ResNet-50~\cite{resnet} configuration. For Swin-Transformer, we use the Swin-B backbone~\cite{SwinTransformer}. We use Swin-B in all final models submitted to the leaderboards and ResNet-50 for validation experiments. The hidden feature dimension is $d=128$.

\vspace{-0.05in}
\paragraph{Pixel decoder.} We adopt the deformable DETR decoder~\cite{deformabledetr} as the multi-scale pixel-wise decoder. Specifically, we use six deformable attention layers to generate feature maps, and the resolutions are the same as Mask2Former~\cite{mask2former}. We have in total $L=3$ layers of feature maps. We add sinusoidal positional and learnable scale-level embedding to the feature maps following~\cite{maskformer}.

\vspace{-0.05in}
\paragraph{Pixel-wise feature propagation.} During feature propagation, we use the simple version of FlowNet~\cite{flownet,flowGuidedFeatureAggregation} pre-trained on the Flying Chairs~\cite{flownet} dataset. The generated flow field is scaled to match the resolution of feature maps with bilinear interpolation.

\vspace{-0.05in}
\paragraph{Object decoder.} The object decoder also has $L=3$ layers. We adopt $N=100$ queries, which are initialized as all-zeros and are learnable embeddings~\cite{mask2former} during training. No dropout~\cite{dropout} is adopted since it would deteriorate the performance of the meta architecture~\cite{mask2former}.

\vspace{-0.05in}
\paragraph{Pixel-wise association.} The thresholds in the pixel-wise association are $\eta_m=0.65$ and $\eta_s=0.80$ on all benchmarks. We found that the P3AFormer model is robust under a wide range of thresholds (see supplementary). The lost tracks are deleted if they do not appear after $n_k=30$ frames.

\vspace{-0.05in}
\paragraph{Training process.} The input image is of shape $1440 \times 800$ for MOT17/MOT20 and $1280\times 384$ for KITTI. Following~\cite{yolox,bytetrack}, we use data augmentation, such as Mosaic~\cite{yolov4} and Mixup~\cite{mixup,automixup}. We use AdamW~\cite{adaw} with an initial learning rate of $6\times 10^{-5}$. We adopt the poly learning rate schedule~\cite{poly} with weight decay $1\times 10^{-4}$. The full training procedure lasts for 200 epochs. The P3AFormer models are all trained with eight Tesla V100 GPUs. The specific configurations of the losses are provided in the supplementary. The run-time analysis of different models is provided in the supplementary.


\setlength{\tabcolsep}{4pt}
\begin{table}[t]
\begin{center}
\caption{Results on the KITTI test split. We show results of two classes ``car" and ``person" on top and bottom splits. The numbers are from~\cite{ObjectPerformance}. We use \textbf{bold} numbers to indicate the best approach of each class.}
\label{tab:KITTITestResults}
\begin{tabular}{l|l|ccccc} \toprule
 Classes & Methods & \multicolumn{1}{c}{HOTA $\uparrow$} & \multicolumn{1}{c}{MOTA $\uparrow$} & \multicolumn{1}{c}{MT $\uparrow$} & \multicolumn{1}{c}{PT $\downarrow$} & \multicolumn{1}{c}{ML $\downarrow$} \\ \midrule
\multirow{9}{*}{Car} & MASS~\cite{MASS} & $68.3$ & $84.6$ & $74.0$ & $23.1$ & $2.9$ \\
 & IMMDP~\cite{rlMOT} & $68.7$ & $82.8$ & $60.3$ & $27.5$ & $12.2$ \\
 & AB3D~\cite{AB3D} & $69.8$ & $83.5$ & $67.1$ & $21.5$ & $11.4$ \\
 & TuSimple~\cite{tusimple} & $71.6$ & $86.3$ & $71.1$ & $22.0$ & $6.9$ \\
 & SMAT~\cite{SMAT} & $71.9$ & $83.6$ & $62.8$ & $31.2$ & $6.0$ \\
 & TrackMPNN~\cite{TrackMPNN} & $72.3$ & $87.3$ & $84.5$ & $13.4$ & \textbf{2.2} \\
 & CenterTrack~\cite{centerTrack} & $73.0$ & $88.8$ & $82.2$ & $15.4$ & $2.5$ \\
 & PermaTrack~\cite{ObjectPerformance} & 78.0 & \textbf{91.3} & 85.7 & 11.7 & 2.6 \\
 & P3AFormer & \textbf{78.4} & 91.2 & \textbf{86.5} & \textbf{10.9} & 2.3 \\ \midrule
\multirow{6}{*}{Person} & AB3D~\cite{AB3D} & 35.6 & 38.9 & 17.2 & 41.6 & 41.2 \\
 & TuSimple~\cite{tusimple} & 45.9 & 57.6 & 30.6 & 44.3 & 25.1 \\
 & TrackMPNN~\cite{TrackMPNN} & 39.4 & 52.1 & 35.1 & 46.1 & 18.9 \\
 & CenterTrack~\cite{centerTrack} & 40.4 & 53.8 & 35.4 & 43.3 & 21.3 \\
 & PermTrack~\cite{ObjectPerformance} & 48.6 & 66.0 & 48.8 & 35.4 & 15.8 \\
 & P3AFormer & \textbf{49.0} & \textbf{67.7} & \textbf{49.1} & \textbf{33.2} & \textbf{14.5} \\
 \bottomrule
\end{tabular}
\end{center}
\end{table}
\setlength{\tabcolsep}{1.4pt}

\begin{figure}[th]
\centering
\includegraphics[width=\textwidth]{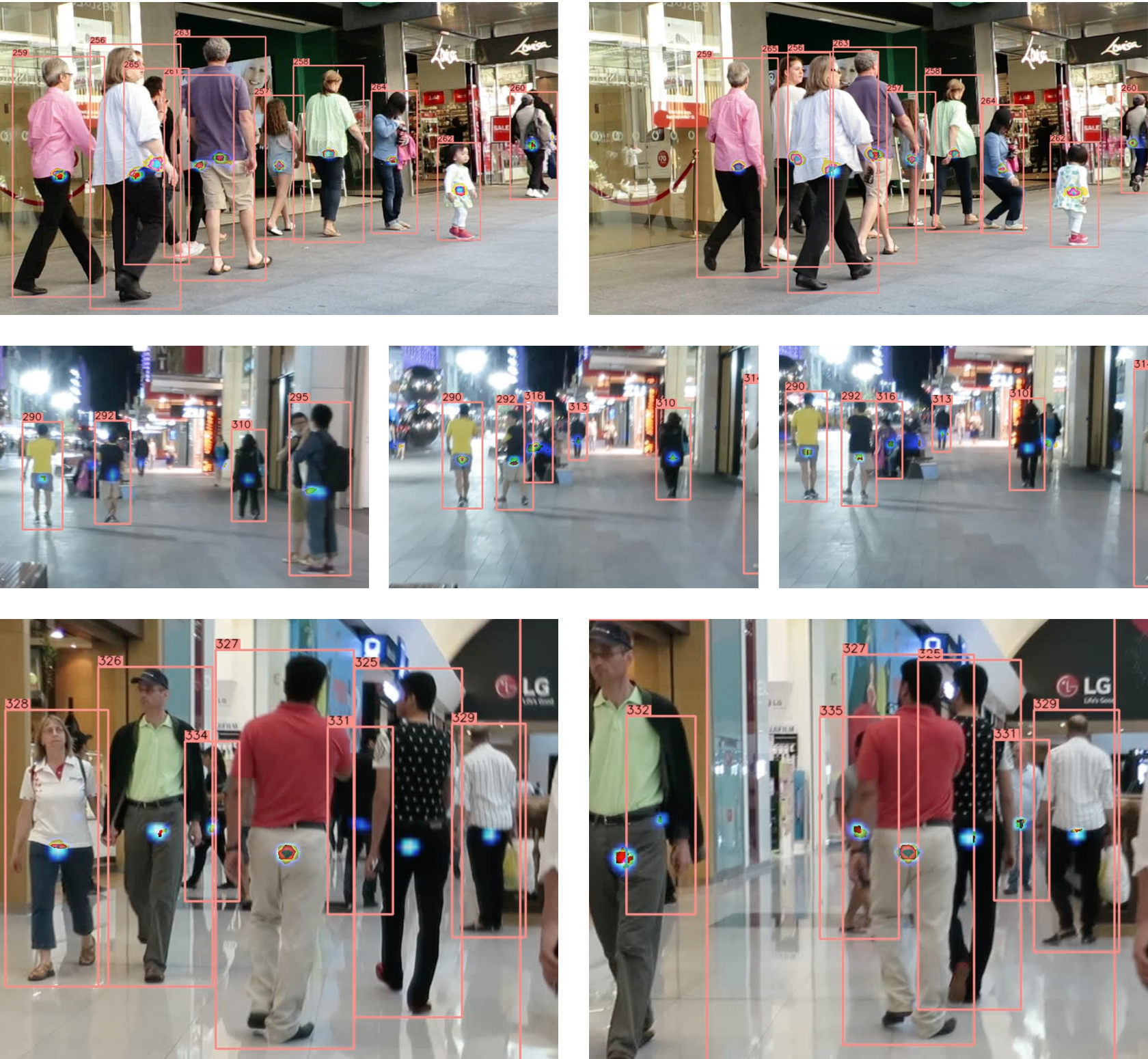}
\caption{Visualization of center heatmaps and tracking results of P3AFormer on \texttt{MOT17-val}. Each row corresponds to one video. The center heatmaps of different objects are put together into one frame and are best viewed on the screen.}
\label{fig:visualization_heatmap}
\end{figure}

\subsection{Comparisons on public benchmarks}
We first compare the P3AFormer model with several baselines on the MOT17 test sets, and the results are presented in Table~\ref{tab:mot17TestResults}. With whistles and bells, P3AFormer outperforms all previous approaches on the two major metrics of MOTA and IDF1. Besides, P3AFormer surpasses the concurrent unpublished transformer-based approaches by a large margin (4.5\% MOTA and 3.0\% IDF1). P3AFormer outperforms the strong unpublished baseline ByteTrack~\cite{bytetrack} while our model differs from theirs. Further, our association procedure does not involve additional training parameters, unlike those of~\cite{Transcenter,GraphMOT,MOTR,centerTrack}.

We also report results on the MOT20 test server in Table~\ref{tab:mot20TestResults}. Again, P3AFormer demonstrates superior performance with whistles and bells. It outperforms SOTA methods \cite{fairmot,coortracker,siammot,SOTMOT} and even the unpublished work~\cite{bytetrack,relationtrack}. Besides, P3AFormer outperforms the concurrent transformer-based work by a large margin (13.1\% MOTA and 17.0\% IDF1). It achieves the best results on this leaderboard.

A comparison between P3AFormer and the baselines on the KITTI dataset is given in Table~\ref{tab:KITTITestResults}. Our work outperforms all baselines on two object classes. Notably, P3AFormer surpasses the strong baseline PermaTrack~\cite{ObjectPerformance} that leverages additional synthetic training data to overcome occlusions. Intriguingly, our P3AFormer does not need those additional training data.

\subsection{Effectiveness of pixel-wise techniques}
\label{sec:ablate_techniques}
We decouple the P3AFormer's pixel-wise techniques and validate the contribution of each part. We use ``Pro." to denote the pixel-wise feature propagation, ``Pre." to denote pixel-wise prediction, and ``Ass." to denote the pixel-wise association. The details of the ablated models are in the supplementary file.

The results are presented in Table~\ref{tab:ablationPixelWise}. We also report the detection mean average precision (mAP~\cite{coco}). The results are much worse when removing all pixel-wise techniques (the first row of Table~\ref{tab:ablationPixelWise}). Compared to the last row, the incomplete system yields 9.2\% mAP, 10.1\% MOTA, and 9.2\% IDF1 lower results. 

When we remove the pixel-wise propagation or pixel-wise prediction (2nd and 3rd rows of Table~\ref{tab:ablationPixelWise}), the results are worse in terms of the detection mAP. Finally, we try different combinations of pixel-wise techniques (4th and 5th rows of Table~\ref{tab:ablationPixelWise}). These combinations improve the tracking performance.

\begin{table}[t]
\begin{center}
\noindent\begin{minipage}[t]{.49\linewidth}
  \caption{Comparison of the pixel-wise techniques on \texttt{MOT17-val}. Please refer to Sec.~\ref{sec:ablate_techniques} for more details.}
  \label{tab:ablationPixelWise}
  \centering
\begin{tabular}{l|ccc}
\toprule
Methods & mAP $\uparrow$& MOTA $\uparrow$ & IDF1 $\uparrow$\\ \midrule
w/o All & 39.1 & 68.3 & 66.8 \\
Pro. & 43.5 & 73.6 & 73.2 \\
Pre. & 42.1 & 72.8 & 74.5 \\
Pre.+Ass. & 41.9 & 71.8 & 74.0 \\
Pro.+Pre. & 48.3 & 69.1 & 72.3 \\
Pro.+Pre.+Ass. & 48.3  & 78.4 & 76.0 \\
\bottomrule
\end{tabular}
 \centering
\end{minipage}\hfill
\begin{minipage}[t]{.49\linewidth}
  \caption{Ablation of the whistles and bells on the \texttt{MOT17-val} (see Sec.~\ref{sec:training_techniques}).}
  \label{tab:ablationTrainingTechniques}
  \centering
\begin{tabular}{l|ccc}
\toprule
Methods & mAP $\uparrow$& MOTA $\uparrow$ & IDF1 $\uparrow$\\ \midrule
w/o All & 46.1 & 71.3 & 72.1 \\
w/o Mask. & 48.0 & 71.4 & 74.7 \\
w/o Mix. & 47.8 & 76.6 & 74.8 \\
w/o Mosiac & 46.7 & 74.0 & 71.9 \\
w/o LQ & 47.9 & 77.6 & 75.1 \\
w/o Bbox & 48.3 & 79.1 & 74.8 \\
with All & 48.3 & 78.4 & 76.0 \\
\bottomrule
\end{tabular}
\centering
\end{minipage}
\hspace{2.3in}

\begin{minipage}[t]{\linewidth}
  \caption{Generalization of pixel-wise techniques to other trackers (refer to Sec.~\ref{sec:generalize_other_trackers}) on the \texttt{MOT17-val}. The validation results of P3AFormer are provided at the bottom for reference.}
  \label{tab:generalizationOther}
  \centering
\begin{tabular}{l|ccccccc}
\toprule
 Methods & \multicolumn{1}{c}{MOTA $\uparrow$} & \multicolumn{1}{c}{IDF1 $\uparrow$} & \multicolumn{1}{c}{MT $\uparrow$} & \multicolumn{1}{c}{ML $\downarrow$} & \multicolumn{1}{c}{FP $\downarrow$} & \multicolumn{1}{c}{FN $\downarrow$} & \multicolumn{1}{c}{IDSW $\downarrow$} \\ \midrule
Tractor~\cite{tractor} & 61.9 & 64.7 & 35.3 & 21.4 & 323 & 42454 & 326 \\
Tractor~\cite{tractor} + Pro. & 63.3 & 67.1 & 37.5 & 20.4 & 310 & 40930 & 279 \\
Tractor~\cite{tractor} + Pro. + Pre.& 64.6 & 69.3 & 38.1 & 17.9 & 287 & 39523 & 238 \\
Tractor~\cite{tractor} + Pro. + Pre. + Ass. & 73.1 & 72.3 & 45.0 & 16.9 & 224 & 30000 & 208 \\
\midrule P3AFormer & 78.9 & 76.3 & 54.3 & 13.6 & 216 & 23462 & 193 \\
\bottomrule
\end{tabular}
\centering
\end{minipage}
\end{center}
\end{table}

\subsection{Influence of training techniques}
\label{sec:training_techniques}

P3AFormer also incorporates several techniques for training transformers. The results are presented in Table~\ref{tab:ablationTrainingTechniques}. We use ``Mask." to represent mask attention -- it is beneficial to detection (0.3 mAP) and association (1.3\% IDF1). We then verify the effect of mixing datasets (CrowdHuman~\cite{crowdhuman}, Cityperson~\cite{Citypersons} and ETHZ~\cite{ethz}) by comparison with only using MOT17 dataset (denoted as ``w/o Mix." in Table~\ref{tab:ablationTrainingTechniques}). It is also clear that using external datasets improves detection and tracking performance. Besides, we notice that using Mosaic augmentation (4th row of Table~\ref{tab:ablationTrainingTechniques}), using learnable query (5th row in Table~\ref{tab:ablationTrainingTechniques}) and connecting all bounding boxes (6th row in Table~\ref{tab:ablationTrainingTechniques}) all slightly improve P3AFormer. 

\subsection{Generalizing to other trackers}
\label{sec:generalize_other_trackers}
Although our pixel-wise techniques are implemented on the transformer structure, one can apply the pixel-wise techniques to other trackers. We consider the tracking-by-detection tracker Tractor~\cite{tractor}, which is based on Faster R-CNN~\cite{fasterrcnn} with a camera motion model and a re-ID network. 

First, we apply pixel-wise feature propagation to Tractor. Second, we change the output shape of faster-RCNN to predict pixel-wise information. After that, we remove the association procedure of the Tractor and replace it with our dense association scheme. More details are included in this generalization experiment in the supplementary file. The results of the above models are presented in Table~\ref{tab:generalizationOther}. It is clear that tracking objects as pixel-wise distributions also improves CNN-based frameworks.

\subsection{Visualization of results}
Visualization of tracking results in comparison to several transformer-based approaches is provided in Figure~\ref{fig:teaser}. The P3AFormer can robustly track small objects without many ID switches. Besides, we provide the visualization of center heatmaps and tracking results of P3AFormer in Figure~\ref{fig:visualization_heatmap}. Even when the objects are heavily occluded, the predicted pixel-wise distribution can provide useful clues to recover the relationship between objects.
\section{Conclusion}
In this paper, we have presented the P3AFormer, which tracks objects as pixel-wise distributions. First, P3AFormer adopts dense feature propagation to build connections through frames. Second, P3AFormer generates multi-level heatmaps to preserve detailed information from the input images. Finally, the P3AFormer exploits pixel-wise predictions during the association procedure, making the association more robust.

P3AFormer obtains state-of-the-art results on three public benchmarks of MOT17, MOT20, and KITTI. P3AFormer demonstrates strong robustness against occlusion and outperforms concurrent transformer-based frameworks significantly. The proposed pixel-wise techniques can be applied to other trackers and may motivate broader exploration of transformers in future work. We will also study the transformer architecture and make it more efficient to satisfy the real-time tracking requirement.

\clearpage
%
%
\bibliographystyle{splncs04}
\bibliography{P3AFormer_reference}
\end{document}